\documentclass{article}

 \usepackage[preprint]{neurips_2026}

\usepackage[utf8]{inputenc} 
\usepackage[T1]{fontenc}    
\usepackage[pagebackref=true,breaklinks=true,colorlinks,citecolor=blue,urlcolor=blue,linkcolor=blue,bookmarks=false]{hyperref}
\usepackage{url}            
\usepackage{booktabs}       
\usepackage{amsfonts}       
\usepackage{nicefrac}       
\usepackage{microtype}      
\usepackage{xcolor}         
\usepackage{graphicx}
\usepackage{booktabs}
\usepackage{amsmath}
\usepackage{multirow}
\usepackage{colortbl}
\usepackage{graphicx}
\definecolor{line-blue}{RGB}{238, 243, 252}
\usepackage{graphicx}
\usepackage{subcaption}
\usepackage{adjustbox}
\usepackage[table]{xcolor}

\title{GeoWeaver: Grounding Visual Tokens with Geometric Evidence before Scene Reasoning}

\author{%
  \textbf{Deshui Miao}$^{1}$ \quad
  \textbf{Xingsen Huang}$^{4}$ \quad
  \textbf{Yameng Gu}$^{1}$ \quad
  \textbf{Xin Li}$^{1,2,3}$\thanks{Corresponding authors.} \\
  \textbf{Haijun Zhang}$^{1,2}$\footnotemark[1] \quad
  \textbf{Ming-Hsuan Yang}$^{5}$ \\
  $^1$Harbin Institute of Technology, Shenzhen \quad
  $^2$Pengcheng Laboratory \\
  $^3$Pazhou Lab (Huangpu) \quad
  $^4$Hainan University \quad
  $^5$University of California at Merced
}

\begin{document}

\maketitle

\begin{abstract}

Spatio-temporal reasoning in vision-language models requires visual representations that preserve physical geometry rather than merely semantic appearance. 
Recent multimodal models incorporate geometric information through structural branches, 3D-aware supervision, reasoning-stage fusion, or long-horizon memory. 
While these approaches demonstrate the importance of geometry for spatial intelligence, they typically treat geometric cues as a shared signal across all visual tokens. 
We note that this overlooks a finer-grained challenge: different visual tokens require different geometric evidence depending on their spatial roles.
To address this limitation, we introduce \textbf{GeoWeaver}, a pre-reasoning geometric grounding framework that treats geometry as a representational prerequisite for spatio-temporal reasoning. 
GeoWeaver constructs a multi-level geometry bank from a frozen geometry encoder and performs token-adaptive geometric evidence allocation, enabling each visual token to retrieve the most relevant geometric abstractions. 
The selected evidence is incorporated into visual tokens via a residual grounding operation prior to language modeling, yielding geometry-grounded representations for downstream reasoning. 
Extensive evaluations on spatial reasoning benchmarks demonstrate that GeoWeaver consistently enhances geometry-aware reasoning while retaining general multimodal capabilities. 
This indicates that geometric information yields the greatest benefit not as a late-fusion auxiliary signal but as a fundamental prerequisite that shapes the representational foundation on which large language models perform reasoning.
All source code and models will be released at \url{https://github.com/yahooo-m/GeoWeaver}.

\end{abstract}

\begin{figure*}[t]
    \centering

    \begin{subfigure}[t]{0.92\textwidth}
        \centering
        \includegraphics[width=\linewidth]{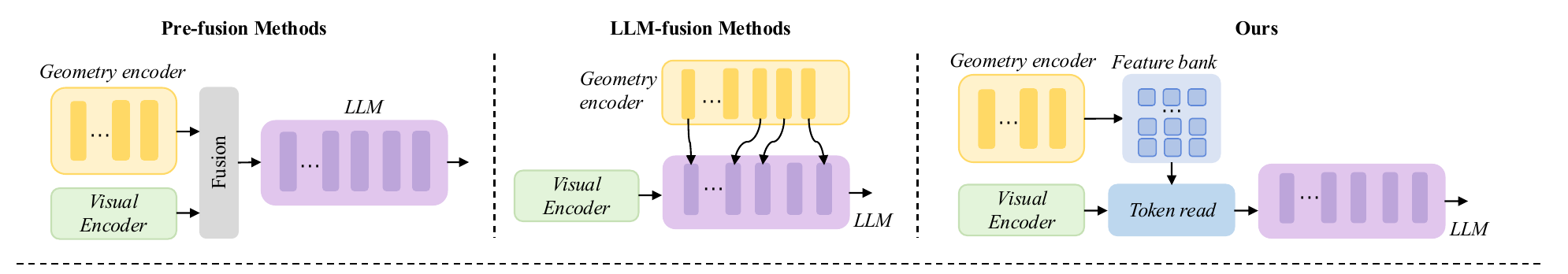}
        \caption{Comparison between pre-fusion methods, LLM-fusion methods, and our pre-reasoning grounding paradigm.}
        \label{fig:overall_scheme}
    \end{subfigure}

    \vspace{0.8em}

    \begin{subfigure}[t]{0.92\textwidth}
        \centering
        \includegraphics[width=\linewidth]{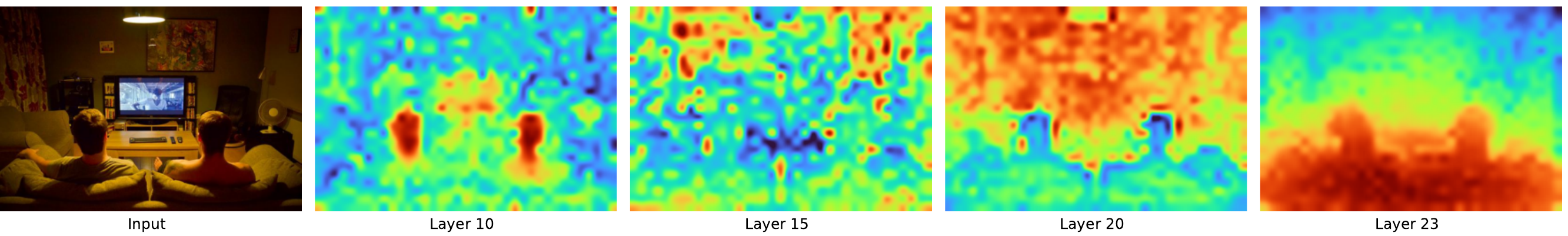}
        \caption{Visualization of VGGT feature responses from different layers.}
        \label{fig:vggt_feature_maps}
    \end{subfigure}

    \caption{
    \textbf{Motivation and paradigm comparison of GeoWeaver.}
    Top: Existing geometry-enhanced VLMs mainly introduce geometry through pre-fusion or LLM-side fusion, while GeoWeaver grounds visual tokens before language reasoning.
    Bottom: VGGT feature maps from different layers exhibit heterogeneous spatial responses, indicating that multi-layer geometry does not provide a uniform signal. This motivates our design of treating geometry features as a multi-level evidence bank.
    }
    \label{fig:motivation_and_scheme}
    \vspace{-7mm}
\end{figure*}

\section{Introduction}

\label{sec:intro}

Understanding and reasoning about physical space is a fundamental capability for intelligent systems that must perceive, communicate, and act in the real world. Recent advances in large vision-language models (VLMs) significantly improve general visual understanding, enabling strong performance in image and video question answering, instruction follow-up, and multimodal reasoning. However, reliable spatio-temporal reasoning remains a major challenge. Tasks involving relative distance estimation, object layout understanding, temporal tracking of spatial relations, or grounding motion and direction in dynamic scenes continue to expose systematic failures that cannot be explained by language reasoning alone.

To address this limitation, recent studies increasingly incorporate spatial information into multimodal models, progressing from perception-side enhancement toward reasoning-oriented spatial modeling. As shown in Figure~\ref{fig:motivation_and_scheme}, early efforts strengthen visual representation by introducing explicit spatial or structural branches, allowing geometry-aware cues to complement semantics-dominated visual features~\cite{wu2025spatial, zheng2025learning}. Building on this direction, representation-level approaches further adopt 3D-aware supervision or geometry-oriented objectives to improve the structural sensitivity of visual tokens, suggesting that spatial reasoning depends not only on input design but also on the geometry encoded in intermediate representations. More recent geometry-enhanced MLLMs investigate how structural cues interact with language reasoning, either by injecting geometric features into the multimodal reasoning stack or by aligning spatial signals with hierarchical reasoning stages~\cite{fan2025vlm,li2025spatialladder}. Beyond single-instance reasoning, long-horizon spatial cognition extends the problem to continuous video streams, where predictive sensing, memory organization, and event-centric modeling support temporally persistent spatial understanding~\cite{yang2025cambrian}. Despite these advances, existing approaches still largely treat geometry as an auxiliary signal introduced at specific stages of the pipeline, rather than as a prerequisite that shapes visual representations before language reasoning begins.

This perspective leaves the representational role of geometry underexplored. We note that spatio-temporal reasoning depends not only on the availability of geometric cues, but also on whether visual tokens are \emph{geometry-grounded} before entering the language model. Without such grounding, the decoder must jointly resolve visual geometry alignment and high-level relational reasoning. This entanglement pushes a perceptual grounding problem into the reasoning stack, reducing the reliability of spatial reasoning and blurring the division between perception and cognition.

Motivated by this observation, we formulate spatio-temporal reasoning as a problem of \emph{pre-reasoning geometric grounding}. Instead of injecting geometry after visual tokens are formed, we use geometry to shape representations before language decoding. Specifically, a frozen geometry encoder constructs a multi-level geometry bank that serves as a reservoir of structural evidence. Each visual token queries this bank and retrieves the geometric abstractions most relevant to its spatial role. The retrieved evidence is then injected into the visual representation, producing geometry-grounded visual tokens for the language model. This formulation shifts visual--geometric alignment from the reasoning stack to the perceptual interface, allowing the decoder to focus on relational and temporal reasoning over geometry-grounded representations.

We instantiate this formulation through a token-adaptive evidence allocation mechanism. The mechanism aligns multi-level geometry tokens with visual tokens, predicts routing weights over geometry layers for each visual token, and applies sparse routing to select compact geometric evidence. As a result, the model avoids uniform exposure to all geometry layers and reduces interference from redundant structural signals. Extensive experiments show that GeoWeaver performs favorably against strong spatial reasoning models across multiple benchmarks while preserving general multimodal capability. Ablation studies further validate the importance of reasoning-relevant geometry layers and sparse evidence allocation, supporting the effectiveness of geometry-grounded visual representations for spatio-temporal reasoning. The main contributions of this work are:
\begin{itemize}
    \item We revisit spatio-temporal reasoning from a representational perspective and identify the lack of geometry-grounded visual tokens as a key limitation of current VLMs.

    \item We propose \textbf{GeoWeaver}, a geometric grounding framework that allocates token-specific geometric evidence from a multi-level geometry bank prior to language decoding.
    
    \item We demonstrate consistent improvements on geometry-sensitive spatial reasoning benchmarks and show through ablations that both geometry-bank construction and evidence allocation are critical to the proposed grounding process.
\end{itemize}

\section{Related Work}
\label{sec:related}

\noindent\textbf{Spatial Intelligence in MLLMs.}
Recent large multimodal models (MLLMs)~\cite{liu2023visual, bai2025qwen3, wang2025internvl3, liu2026enhancing} show strong general-purpose visual understanding, but spatial reasoning in physical environments remains challenging. Recent benchmarks, including VSI-Bench~\cite{yang2025thinking}, SPAR~\cite{zhang2025flatland}, BLINK~\cite{fu2024blink}, and Cambrian-S~\cite{yang2025cambrian}, evaluate this limitation through tasks involving depth, viewpoint, topology, and spatiotemporal relations~\cite{openai_gpt5_systemcard, gemini_3_pro_systemcard, zhou2025vlm4d}. These studies indicate that spatial intelligence is not merely a difficult subset of generic multimodal reasoning. Instead, it requires visual representations that preserve geometry-sensitive evidence beyond semantics-dominated appearance cues. Our work follows this diagnosis, but focuses on the representation before reasoning: whether visual tokens are geometrically grounded when they enter the language model.

\noindent\textbf{Data Scaling and Spatial Reasoning Supervision.}
Much effort improves spatial reasoning through data construction and specialized optimization. SpatialVLM~\cite{chen2024spatialvlm}, SPAR~\cite{zhang2025flatland}, and Cambrian-S~\cite{yang2025cambrian} enlarge the supervision space with synthetic 3D scenes, physically grounded tasks, and long-horizon spatial cognition settings. Furthermore, VST~\cite{vst2025}, SpaceR~\cite{ouyang2025spacer}, MindCube~\cite{yin2025spatial(mindcube)}, SpatialLadder~\cite{li2025spatialladder}, and Spatial-SSRL~\cite{liu2025spatial} exploit structured tuning, reinforcement learning, and verifiable rewards to strengthen spatial reasoning behaviors. SenseNova-SI~\cite{cai2025scaling} further suggests that scaling text-based chain-of-thought alone brings limited gains, highlighting that spatial reasoning depends on more than language-side deliberation. These methods provide stronger supervision and optimization signals, while our work addresses a complementary question: how the visual representation should be organized before such reasoning takes place.

\noindent\textbf{Geometry-Augmented Multimodal Modeling.}
A closely related direction incorporates explicit geometric priors into MLLMs. Some methods introduce external 3D inputs, such as point clouds or RGB-D signals, to connect language models with spatial structure~\cite{du2024embspatial, zhu2025llava, zheng2025video}. More recently, feed-forward geometry encoders, including DUST3R~\cite{wang2024dust3r, leroy2024grounding}, CUT3R~\cite{wang2025continuous}, and VGGT~\cite{wang2025vggt}, provide dense geometric estimates from multi-view images or videos. Building on these encoders, Spatial-MLLM~\cite{wu2025spatial}, VG-LLM~\cite{zheng2025learning}, VLM-3R~\cite{fan2025vlm, fan2026vlm3r}, SSR~\cite{liu2025ssr}, and related models~\cite{li2026thinking} introduce geometry through structural branches, patch-level fusion, cross-attention, or rationale-guided integration. These approaches demonstrate the value of explicit geometry, but they primarily introduce geometric information as an additional signal after or alongside visual representations. In contrast, GeoWeaver uses geometry to ground visual tokens before language decoding, making geometric evidence part of the representation on which reasoning operates.

\noindent\textbf{Multimodal Chain-of-Thought and Visual Intermediate Reasoning.}
Recent unified multimodal architectures~\cite{team2024chameleon, deng2025bagel, jiao2025unitoken} also motivate a line of work that externalizes intermediate reasoning through visual generation or transformation. For example, some methods render maze trajectories~\cite{wu2024mind}, imagine temporal transitions~\cite{li2025imagine}, or draw intermediate 2D geometric cues~\cite{li2025zebra, gu2025thinkmorph}. Such visual intermediates help when the required transformation is difficult to express purely in text. However, spatio-temporal reasoning in real 3D scenes also requires dense geometric structure, viewpoint consistency, and continuous physical relations to remain available inside the representation. GeoWeaver does not externalize reasoning into generated visual artifacts; instead, it improves the visual tokens consumed by the language model through pre-reasoning geometric grounding.

\section{Method}
\label{sec:method}
Motivated by the observation that spatio-temporal reasoning requires geometrically grounded visual representations before language decoding, GeoWeaver adopts a ground-then-reason design that uses geometry to prepare visual tokens rather than treating them as a late fusion signal. This separates low-level visual--geometric alignment from high-level relational and temporal reasoning, allowing the language model to operate on geometry-ready representations. 
We first introduce the semantic and geometric feature extraction branches, then describe token-adaptive geometric grounding, and finally present language reasoning over grounded visual tokens and the training objective.

\subsection{Overview}

Given a sequence of input frames
$\{\mathbf{I}_k \in \mathbb{R}^{H \times W \times 3}\}_{k=1}^{K}$
and a language instruction $q$, our framework consists of three stages:
(1) a semantic visual branch extracts visual tokens from the base VLM,
(2) a frozen geometry branch constructs a multi-level geometry bank from the same frames, and
(3) a token-adaptive grounding stage allocates geometric evidence to visual tokens before language decoding.

\begin{figure}[t]
    \centering
    \includegraphics[width=\linewidth]{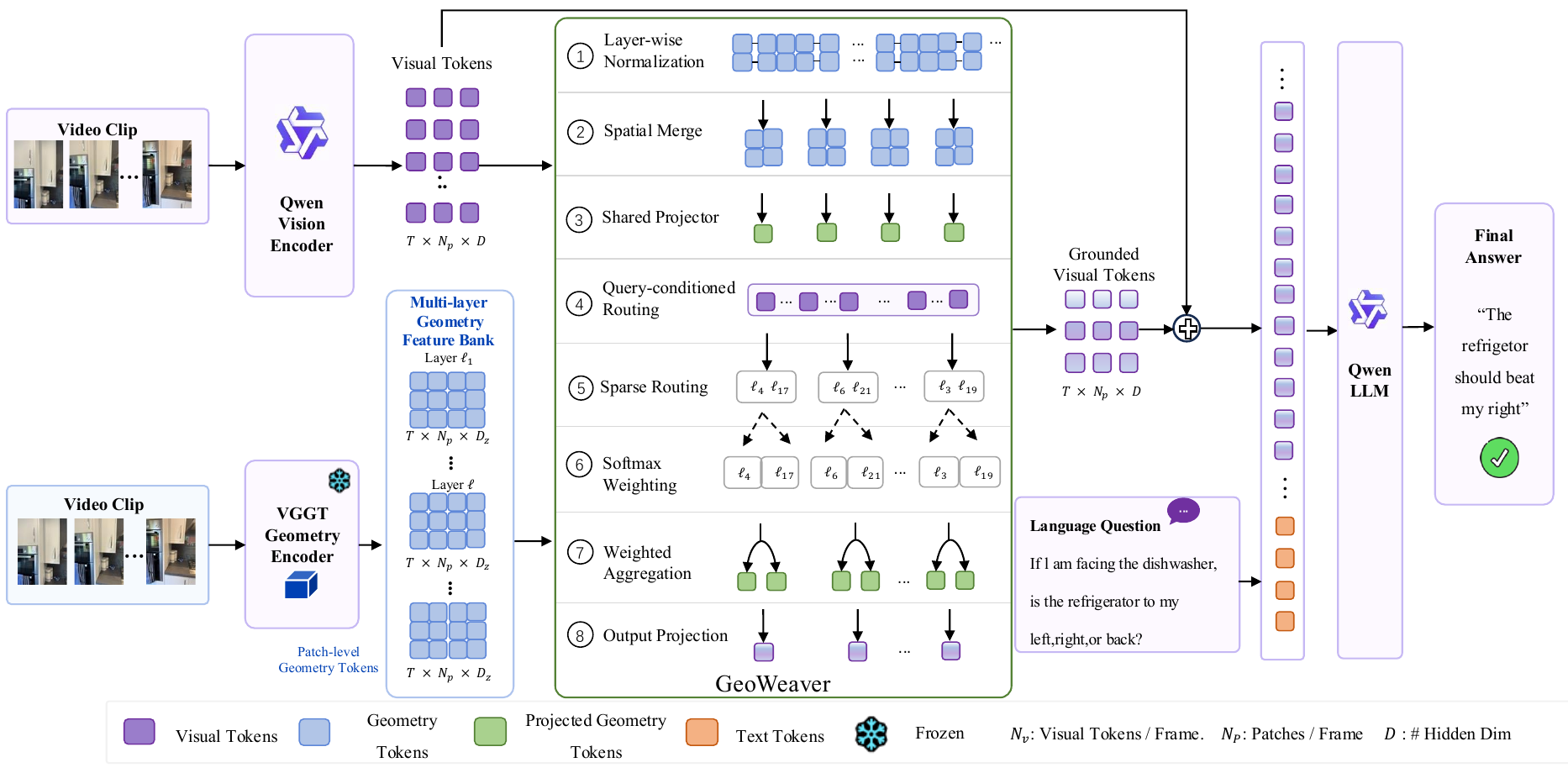}
    \vspace{-2mm}
     \caption{
\textbf{Overview of GeoWeaver.}
GeoWeaver treats geometry as a representational prerequisite rather than a late fusion signal. A frozen VGGT encoder provides a multi-layer geometry bank, from which each visual token adaptively retrieves sparse geometric evidence via query-conditioned compact geometric grounding before entering the Qwen LLM. This pre-reasoning grounding process converts semantic visual tokens into geometry-grounded visual tokens for spatio-temporal reasoning.
}
    \label{fig:overall}
    \vspace{-4mm}
\end{figure}
\noindent \textbf{Visual branch.}
We build our method on top of a pretrained multimodal LLM. In our implementation, we use a Qwen3.5 VLM backbone, although the framework is model-agnostic. Given $K$ input frames, the native vision encoder produces patch tokens, which are further processed by the model's built-in token merger. Let
\begin{equation}
\mathbf{V} \in \mathbb{R}^{(K N_p) \times D}
\label{eq:visual_tokens}
\end{equation}
denote the resulting visual tokens, where $N_p$ is the number of merged spatial tokens per frame and $D$ is the hidden dimension expected by the language model.

\noindent \textbf{Geometry branch.}
In parallel, we use a frozen geometry encoder to extract explicit geometric priors from the same input frames. We adopt VGGT~\cite{wang2025vggt} as the geometry encoder. Rather than using only its final representation, we collect a contiguous set of deeper hidden states to form a \emph{multi-level geometry bank}. Specifically, for each selected layer $l \in \mathcal{S}$, we take the corresponding patch-token hidden state, discard non-patch tokens such as camera or register tokens, and obtain
\begin{equation}
\mathbf{Z}^{(l)} \in \mathbb{R}^{(K N_v) \times D_{\text{geo}}},
\qquad l \in \mathcal{S},
\label{eq:geo_bank_raw}
\end{equation}
where $N_v$ is the number of raw geometry tokens per frame and $D_{\text{geo}}$ is the geometry feature dimension. In practice, we use the latter half of the geometry encoder layers (layers 12--23 in a 24-layer VGGT), which provide rich structural abstractions while preserving spatial detail.

\subsection{Token-Adaptive Geometric Grounding}

The core of our framework is a token-adaptive grounding process that injects geometric evidence into visual tokens before language decoding. The process consists of four steps: per-layer normalization, spatial alignment, shared projection, and query-conditioned sparse routing.

\noindent \textbf{Per-layer normalization and spatial alignment.}
As shown in Fig.~\ref{fig:motivation_and_scheme}, geometry features from different layers exhibit distinct distributions and resolutions. We therefore normalize each selected geometry layer independently:
\begin{equation}
\hat{\mathbf{Z}}^{(l)} = \mathrm{LN}\!\left(\mathbf{Z}^{(l)}\right),
\qquad l \in \mathcal{S}.
\label{eq:layernorm}
\end{equation}
Because the geometry branch and base VLM operate at different spatial scales, direct token-wise grounding is not feasible without alignment. To match the visual token resolution, we apply a $2 \times 2$ spatial merge to the normalized geometry tokens, followed by a shared projector:
\begin{equation}
\mathbf{G}^{(l)} = \phi\!\left(\mathrm{Merge}_{2\times2}\!\left(\hat{\mathbf{Z}}^{(l)}\right)\right),
\qquad
\mathbf{G}^{(l)} \in \mathbb{R}^{(K N_p) \times D},
\label{eq:geo_align}
\end{equation}
where $\phi(\cdot)$ is a lightweight MLP shared across all geometry layers. After this step, all geometry layers are aligned to the same token resolution and hidden dimension as the visual tokens in Eq.~\ref{eq:visual_tokens}.

\noindent \textbf{Query-conditioned geometric evidence allocation.}
A key question is which geometric abstraction should be assigned to each visual token. Instead of exposing all tokens uniformly to every geometry layer, we let each visual token query the multi-level geometry bank. For the $i$-th visual token $\mathbf{v}_i \in \mathbb{R}^{D}$, a lightweight router predicts its preference over the selected geometry layers:
\begin{equation}
\mathbf{r}_i = f_{\text{router}}(\mathbf{v}_i),
\qquad
\mathbf{r}_i \in \mathbb{R}^{|\mathcal{S}|}.
\label{eq:router}
\end{equation}
This design reflects our hypothesis that different visual regions require different geometric abstractions: local boundaries, depth discontinuities, and occlusion patterns need not depend on the same evidence as global layout or relational structure.

\noindent \textbf{Compact geometric grounding.}
Not all geometry layers provide equally useful evidence for a given visual token. Aggregating the entire geometry bank introduces both relevant and weakly related structural cues, diluting token-specific grounding. We therefore adopt a compact evidence selection strategy that retains only the geometry layers with the highest token-conditioned evidence scores. Given routing logits $\mathbf{r}_i$, we compute layer weights as
\begin{equation}
\boldsymbol{\alpha}_i
=
\mathrm{Softmax}\!\left(\mathcal{S} _{TopK}(\mathbf{r}_i, K)\right),
\label{eq:topk}
\end{equation}
where $\mathcal{S} _{TopK}(\cdot, K)$ keeps the $K$ most relevant geometry layers and masks the remaining entries before normalization.

Given the aligned geometry bank $\{\mathbf{G}^{(l)}\}_{l\in\mathcal{S}}$, the geometric evidence assigned to token $i$ is
\begin{equation}
\mathbf{g}_i
=
\sum_{l \in \mathcal{S}}
\alpha_i^{(l)} \mathbf{G}^{(l)}_i,
\label{eq:agg}
\end{equation}
where $\mathbf{G}^{(l)}_i$ denotes the geometry token aligned with $\mathbf{v}_i$ in layer $l$. This operation aggregates multi-layer geometry into compact token-specific evidence for residual grounding.



\noindent \textbf{Residual grounding before reasoning.}
Finally, we inject the allocated geometric evidence into each visual token through a residual grounding operation:
\begin{equation}
\mathbf{v}_i' = \mathbf{v}_i + \mathbf{W}_o \mathbf{g}_i,
\label{eq:residual_ground}
\end{equation}
where $\mathbf{W}_o \in \mathbb{R}^{D \times D}$ is a projection matrix. We initialize $\mathbf{W}_o$ to zero so that the grounding branch begins as an identity mapping and gradually learns to reshape the visual representation during finetuning. Collecting all grounded visual tokens yields
\begin{equation}
\mathbf{V}' = [\mathbf{v}_1'; \ldots; \mathbf{v}_{K N_p}'] \in \mathbb{R}^{(K N_p) \times D}.
\label{eq:grounded_visual_tokens}
\end{equation}

This step distinguishes our framework from prior geometry-enhanced MLLMs. Rather than injecting geometry into the language model as an auxiliary signal, we use geometry to transform the representations over which the language model reasons. The result is a set of \emph{geometry-grounded visual tokens} better suited for spatio-temporal reasoning.

\subsection{Language Reasoning over Geometry-Grounded Tokens}

After grounding, the visual tokens are concatenated with the text tokens derived from the instruction:
\begin{equation}
\mathbf{H}_0 = [\mathbf{V}'; \mathbf{T}],
\label{eq:multimodal_input}
\end{equation}
where $\mathbf{T} \in \mathbb{R}^{M \times D}$ denotes the text-token embeddings. The resulting multimodal sequence is processed by the LLM decoder:
\begin{equation}
\mathbf{H}^{\text{llm}}_L
=
f^{\text{llm}}_L\!\Big(
f^{\text{llm}}_{L-1}\!\big(
\cdots f^{\text{llm}}_1(\mathbf{H}_0)
\big)
\Big),
\label{eq:llm}
\end{equation}
where $f^{\text{llm}}_j(\cdot)$ denotes the $j$-th decoder block and $\mathbf{H}^{\text{llm}}_L$ the final hidden states. Because grounding is completed before decoding, the LLM no longer needs to infer visual--geometric alignment during reasoning and can instead focus on higher-level relational and temporal reasoning over grounded representations.

\subsection{Optimization}

We train the model using the standard next-token prediction objective~\cite{Qwen3-VL}:
\begin{equation}
\mathcal{L}_{\text{ce}}(\theta)
=
-\sum_{t=1}^{|o|}
\log P_{\theta}\!\left(
o_t \mid o_{<t}, q, \{\mathbf{I}_k\}_{k=1}^{K}
\right),
\label{eq:loss}
\end{equation}
where $o_t$ denotes the $t$-th target answer token. During instruction tuning, we keep the geometry encoder frozen and jointly optimize the grounding module with the base multimodal model. In practice, we apply parameter-efficient finetuning to the language model while training the grounding layers with full parameters. No auxiliary geometric reconstruction or contrastive objectives are introduced; instead, the model learns to allocate and exploit geometric evidence purely through end-to-end supervision on downstream spatio-temporal reasoning tasks.

\section{Experiments}
\label{sec:experiments}

\begin{table*}[tp]
\caption{
    \textbf{Comparison on VSI-Bench~\cite{yang2025thinking}.}
    Numerical answers are evaluated by MRA and multiple-choice answers by accuracy.
    All results follow the EASI protocol~\cite{easi2025}.
    Best and second-best open-source results are \textbf{bolded} and \underline{underlined}, respectively.
}
\centering
\resizebox{.97\textwidth}{!}{%
\begin{tabular}{l|c|cccc|cccc}
    \toprule
    \multirow{2}{*}[-0.6ex]{Methods}
    & \multirow{2}{*}[-0.6ex]{Avg.}
    & \multicolumn{4}{c|}{\cellcolor{orange!10}Numerical Answer}
    & \multicolumn{4}{c}{\cellcolor{yellow!10}Multiple-Choice Answer} \\
    \cmidrule(lr){3-6}\cmidrule(lr){7-10}
    & & Obj. Count & Abs. Dist. & Obj. Size & Room Size
    & Rel. Dist. & Rel. Dir. & Route Plan & Appr. Order \\
    \midrule
    Human & 79.2 & 94.3 & 47.0 & 60.4 & 45.9 & 94.7 & 95.8 & 95.8 & 100.0 \\
    Random Choice(Frequency) & 34.0 & 62.1 & 32.0 & 29.9 & 33.1 & 25.1 & 47.9 & 28.4 & 25.2 \\
    
    \midrule
    \rowcolor{black!9}\multicolumn{10}{l}{\textbf{Proprietary Models}} \\
    Seed-1.6~\cite{seed2025seed1_5vl} & 49.9 & 43.5 & 34.4 & 66.1 & 52.8 & 55.1 & 35.7 & 44.3 & 68.0 \\
    Gemini-3-Pro~\cite{gemini_3_pro_systemcard} & 52.5 & 38.0 & 37.8 & 72.7 & 44.1 & 59.9 & 55.7 & 45.9 & 66.0 \\
    Grok-4~\cite{grok4_xai_2025} & 47.9 & 37.2 & 33.0 & 60.8 & 45.4 & 53.1 & 39.7 & 47.4 & 66.8 \\
    GPT-5~\cite{openai_gpt5_systemcard} & 55.0 & 53.3 & 34.5 & 73.3 & 47.5 & 63.7 & 48.7 & 50.3 & 68.9 \\
    
    \midrule
    \rowcolor{black!9}\multicolumn{10}{l}{\textbf{Open-source General Models}} \\
    Bagel-7B-MoT~\cite{deng2025bagel} & 31.4 & 30.1 & 29.2 & 35.5 & 25.8 & 34.9 & 41.4 & 30.4 & 24.1 \\
    Qwen2.5-VL-7B-Instruct~\cite{Qwen2.5-VL} & 32.3 & 32.9 & 18.2 & 43.9 & 31.7 & 38.0 & 37.4 & 28.4 & 28.0 \\
    Qwen3-VL-8B-Instruct~\cite{Qwen3-VL} & 57.9 & 67.6 & 47.0 & 76.3 & 61.9 & 58.0 & 51.0 & 35.0 & 66.3 \\
    InternVL3-8B~\cite{zhu2025internvl3} & 42.1 & 66.0 & 34.9 & 43.6 & 47.5 & 48.0 & 39.3 & 26.3 & 31.4 \\
    
    \midrule
    \rowcolor{black!9}\multicolumn{10}{l}{\textbf{Open-source Spatial Intelligence Models}} \\
    MindCube-3B~\cite{yin2025spatial(mindcube)} & 17.2 & 12.9 & 22.8 & 4.3 & 23.5 & 20.3 & 15.7 & 16.0 & 22.5 \\
    SpatialLadder-3B~\cite{li2025spatialladder} & 44.9 & 62.2 & 35.4 & 62.0 & 41.4 & 45.6 & 46.5 & 27.3 & 38.5 \\
    Spatial-MLLM-4B~\cite{wu2025spatial} & 46.3 & 66.7 & 38.1 & 63.6 & 35.5 & 40.4 & 48.2 & 33.0 & 44.3 \\
    SpaceR-7B~\cite{ouyang2025spacer} & 41.5 & 44.5 & 24.7 & 53.5 & 37.3 & 41.9 & 46.1 & 29.3 & 54.8 \\
    VLM-3R-7B~\cite{fan2025vlm} & 60.9 & 70.2 & 49.4 & 69.2 & 67.1 & 65.4 & 80.5 & 45.4 & 40.1 \\
    ViLaSR-7B~\cite{wu2025reinforcing(vilasr)} & 44.6 & 58.1 & 33.9 & 61.4 & 28.9 & 45.1 & 46.5 & 29.9 & 53.2 \\
    VST-7B-SFT~\cite{vst2025} & 55.5 & 68.8 & 37.3 & 74.5 & 62.2 & 55.2 & 48.7 & 41.8 & 55.5 \\
    Cambrian-S-7B~\cite{yang2025cambrian} & 62.9 & 68.2 & 45.8 & 72.5 & 67.6 & 66.8 & 69.6 & 39.2 & 73.8 \\
    SpatialStack-5B~\cite{zhang2026spatialstack} & \underline{67.5} & \underline{71.0} & \underline{55.6} & 69.1 & \underline{68.2} & 67.3 & \underline{84.1} & 41.2 & \underline{83.5} \\
    
    \midrule
    \rowcolor{line-blue}\multicolumn{10}{l}{\textbf{Ours}} \\
    \textbf{GeoWeaver-5B}
    & \underline{72.8} (+5.3)
    & \underline{83.2} (+12.2)
    & \textbf{61.8} (+6.2)
    & \underline{76.8} (+7.7)
    & 62.4
    & \underline{72.8} (+5.5)
    & 83.5
    & \textbf{56.3} (+15.1)
    & \textbf{84.7} (+1.2) \\
    
    \textbf{GeoWeaver-10B}
    & \textbf{75.05} (+7.6)
    & \textbf{84.3} (+13.3)
    & \underline{58.8} (+3.2)
    & \textbf{83.8} (+14.7)
    & \textbf{68.8} (+0.6)
    & \textbf{73.5} (+6.2)
    & \textbf{92.0} (+7.9)
    & \underline{55.0} (+13.8)
    & \underline{84.3} (+0.8) \\
    \bottomrule
\end{tabular}%
}
\label{tab:vsi-bench}
\vspace{-5mm}
\end{table*}

\begin{table*}[tp]
\centering
\caption{
\textbf{Comparison on ReVSI~\cite{zhang2026revsi}.}
Baseline results follow the native inference frame settings reported by ReVSI, while GeoWeaver is evaluated under different frame budgets. All scores are reported as accuracy (\%).  The best and second-best results are \textbf{bolded} and \underline{underlined}, respectively.
}
\vspace{-2mm}
\resizebox{0.98\textwidth}{!}{%
\begin{tabular}{l|c|c|cccc|ccc}
    \toprule
    \multirow{2}{*}[-0.6ex]{Methods}
    & \multirow{2}{*}[-0.6ex]{Frames}
    & \multirow{2}{*}[-0.6ex]{Avg.}
    & \multicolumn{4}{c|}{\cellcolor{orange!10}Numerical Question}
    & \multicolumn{3}{c}{\cellcolor{yellow!10}Multiple-Choice Question} \\
    \cmidrule(lr){4-7}\cmidrule(lr){8-10}
    & & & Obj. Count & Abs. Dist. & Obj. Size & Room Size
    & Rel. Dist. & Rel. Dir. & Route Plan \\
    \midrule
    Cambrian-S-7B~\cite{yang2025cambrian}
    & 128 & 49.1
    & \textbf{48.4} & 60.5 & 65.5 & 46.7
    & 37.1 & 48.5 & 37.0 \\

    VST-7B-SFT~\cite{vst2025}
    & 4 FPS & 46.4
    & 35.4 & 52.6 & 67.9 & 47.2
    & 49.2 & 36.9 & 35.4 \\

    SpaceR-7B~\cite{ouyang2025spacer}
    & 32 & 30.5
    & 30.7 & 34.5 & 52.0 & 18.6
    & 22.8 & 34.5 & 20.2 \\

    Spatial-MLLM-4B-135k~\cite{wu2025spatial}
    & 16 & 40.5
    & 40.7 & 45.3 & 46.8 & --
    & 32.3 & 37.4 & -- \\

    Spatial-MLLM-4B-820k~\cite{wu2025spatial}
    & 16 & 40.9
    & 41.5 & 40.0 & 53.1 & --
    & 30.7 & 39.2 & -- \\

    VLM-3R-7B~\cite{fan2025vlm}
    & 32 & \underline{50.1}
    & 41.6 & 61.6 & 64.8 & 52.5
    & 46.5 & 49.5 & 34.1 \\

        \midrule
    \rowcolor{line-blue}\multicolumn{10}{l}{\textbf{Ours}} \\

    \multirow{3}{*}{\textbf{GeoWeaver-5B}}
    & \cellcolor{line-blue}16 
    & \cellcolor{line-blue}54.2 {\small (+4.1)}
    & \cellcolor{line-blue}38.6 
    & \cellcolor{line-blue}61.1 
    & \cellcolor{line-blue}62.8 
    & \cellcolor{line-blue}--
    & \cellcolor{line-blue}61.4 
    & \cellcolor{line-blue}47.0 
    & \cellcolor{line-blue}-- \\

    & \cellcolor{line-blue}32 
    & \cellcolor{line-blue}54.9 {\small (+4.8)}
    & \cellcolor{line-blue}38.0 
    & \cellcolor{line-blue}63.4 
    & \cellcolor{line-blue}63.6 
    & \cellcolor{line-blue}\textbf{54.2}
    & \cellcolor{line-blue}\textbf{65.3} 
    & \cellcolor{line-blue}46.9 
    & \cellcolor{line-blue}53.1 \\

    & \cellcolor{line-blue}All 
    & \cellcolor{line-blue}54.1 {\small (+4.0)}
    & \cellcolor{line-blue}38.6 
    & \cellcolor{line-blue}61.3 
    & \cellcolor{line-blue}62.2 
    & \cellcolor{line-blue}53.7
    & \cellcolor{line-blue}63.8 
    & \cellcolor{line-blue}46.1 
    & \cellcolor{line-blue}53.1 \\
    \midrule

    \multirow{3}{*}{\textbf{GeoWeaver-10B}}
    & \cellcolor{line-blue}16 
    & \cellcolor{line-blue}57.4 {\small (+7.3)}
    & \cellcolor{line-blue}40.5 
    & \cellcolor{line-blue}64.5 
    & \cellcolor{line-blue}69.3 
    & \cellcolor{line-blue}--
    & \cellcolor{line-blue}62.1 
    & \cellcolor{line-blue}50.5 
    & \cellcolor{line-blue}-- \\

    & \cellcolor{line-blue}32 
    & \cellcolor{line-blue}\textbf{57.5} {\small (\textbf{+7.4})}
    & \cellcolor{line-blue}42.6 
    & \cellcolor{line-blue}\textbf{66.8} 
    & \cellcolor{line-blue}\textbf{69.8} 
    & \cellcolor{line-blue}50.9
    & \cellcolor{line-blue}64.4 
    & \cellcolor{line-blue}\textbf{52.4} 
    & \cellcolor{line-blue}\textbf{55.4} \\

    & \cellcolor{line-blue}All 
    & \cellcolor{line-blue}57.3 {\small (+7.2)}
    & \cellcolor{line-blue}43.2 
    & \cellcolor{line-blue}66.5 
    & \cellcolor{line-blue}69.7 
    & \cellcolor{line-blue}50.7
    & \cellcolor{line-blue}65.0 
    & \cellcolor{line-blue}51.7 
    & \cellcolor{line-blue}54.7 \\
    \bottomrule
\end{tabular}%
}

\label{tab:revsi_compare}
\vspace{-2mm}
\end{table*}
\subsection{Experimental Setup}

\noindent \textbf{Benchmarks.}
We evaluate GeoWeaver on both specialized spatial reasoning benchmarks and general multimodal evaluation sets. For spatial reasoning, we report results on \textbf{VSI-Bench}~\cite{yang2025thinking}, \textbf{ReVSI}~\cite{zhang2026revsi}, \textbf{SPAR-Bench}~\cite{zhang2025flatland}, \textbf{ViewSpatial}~\cite{li2025viewspatial}, \textbf{BLINK}~\cite{fu2024blink}, \textbf{3DSRBench}~\cite{ma20253dsrbench}, and \textbf{EmbSpatial}~\cite{du2024embspatial}. Among them, VSI-Bench provides a fine-grained diagnosis over eight subcategories, including four numerical-answer tasks (\emph{Object Count}, \emph{Absolute Distance}, \emph{Object Size}, \emph{Room Size}) and four multiple-choice tasks (\emph{Relative Distance}, \emph{Relative Direction}, \emph{Route Plan}, \emph{Approaching Order}). To examine whether the proposed grounding mechanism preserves general multimodal ability, we further evaluate on \textbf{MMBench}~\cite{liu2024mmbench}, \textbf{Video-MME}~\cite{fu2025video}, \textbf{BLINK}, and \textbf{TempCompass}~\cite{liu2024tempcompass}.


\noindent \textbf{Metrics.}
For VSI-Bench, we follow the EASI protocol and report \emph{MRA} for numerical-answer questions and \emph{accuracy} for multiple-choice questions. For the remaining benchmarks, we report the standard score used by each benchmark.

\noindent \textbf{Implementation Details.}
We instantiate GeoWeaver on Qwen3.5~\cite{Qwen3-VL} multimodal LLMs with both 4B- and 9B-scale backbones, using the native vision encoder of each backbone to extract semantic visual tokens. For geometric grounding, we use a pretrained VGGT encoder as a frozen geometry branch.
Each selected geometry layer is aligned to the visual-token resolution through a $2\times2$ spatial merge and projected into the LLM hidden space by a shared MLP. The grounding stage assigns compact geometric evidence to each visual token through token-conditioned layer selection, with two geometry layers selected by default. The selected evidence is injected into visual tokens through residual grounding, where the output projection is zero-initialized for stable finetuning.
During training, the VGGT encoder remains frozen, while the grounding layers are fully optimized. For the base multimodal LLM, we adopt LoRA-based parameter-efficient finetuning for 9B configurations and full training for 4B MLLM.  

\subsection{Main Results}

\begin{table*}[t]
    \centering
    \caption{\textbf{Spatial ability evaluation.} Comparison with existing methods on spatial reasoning tasks.}
    \vspace{-2mm}
    \setlength{\tabcolsep}{2pt}
    \resizebox{\textwidth}{!}{%
    \begin{tabular}{l|cccccc|c}
        \toprule
        \rowcolor{orange!10}
        Methods & VSI-Bench & SPAR-Bench & ViewSpatial & \hspace{.2em} BLINK \hspace{.2em} & 3DSRBench & EmbSpatial & Overall \\
        \midrule
        VLM-3R-7B~\cite{fan2025vlm} & 60.7 & 42.4 & 40.5 & 52.3 & 51.5 & 68.2 & 52.6 \\
        MindCube-3B~\cite{yin2025spatial(mindcube)} & 17.2 & 20.8 & 24.1 & 35.1 & 2.8 & 37.0 & 22.8 \\
        SpatialLadder-3B~\cite{li2025spatialladder} & 44.9 & 32.9 & 39.9 & 43.0 & 42.8 & 58.2 & 43.6 \\
        Spatial-MLLM-4B~\cite{wu2025spatial} & 46.3 & 35.3 & 34.7 & 40.5 & 36.2 & 50.0 & 40.5 \\
        SpaceR-7B~\cite{ouyang2025spacer} & 41.6 & 34.2 & 35.9 & 49.6 & 40.5 & 66.9 & 44.8 \\
        ViLaSR-7B~\cite{wu2025reinforcing(vilasr)} & 44.6 & 37.4 & 35.7 & 51.4 & 46.6 & 67.3 & 47.2 \\
        Cambrian-S-7B~\cite{yang2025cambrian} & 62.9 & 37.9 & 41.3 & 37.9 & 45.0 & 72.8 & 49.6 \\
        \midrule
        \rowcolor{line-blue} 
        \textbf{GeoWeaver-5B} & \textbf{72.8} {\small (+9.9)} & \textbf{69.2} {\small (+26.8)} & \textbf{52.8} {\small (+11.5)} & \textbf{58.1} {\small (+5.8)} & \textbf{58.6} {\small (+7.1)} & \textbf{77.5} {\small (+4.7)} & \textbf{64.8} {\small (+12.2)} \\
        \bottomrule
    \end{tabular}%
    }
    
    \label{tab:spatial_reasoning_bench}
    \vspace{-4mm}
\end{table*}

\begin{table*}[tp]
    \caption{\textbf{General Capabilities Evaluation.} Our method maintains robust general multimodal and spatio-temporal reasoning capabilities, demonstrating no catastrophic forgetting.}
    \centering
    \resizebox{0.98\linewidth}{!}{%
        \renewcommand{\arraystretch}{0.85}
        \setlength{\tabcolsep}{5pt}
        \begin{tabular}{l|cccc|c}
            \toprule
            \rowcolor{orange!10}
            Method & MMBench & Video-MME & BLINK & TempCompass & Overall \\
            \midrule
            Qwen3.5-4B & 83.25 & 62.44 & \textbf{61.12} & 66.84 & 68.41 \\
            SpatialStack-5B~\cite{zhang2026spatialstack} & \textbf{83.42} & 63.74 & 55.46 & 69.37 & 68.00 \\
            \midrule
           \rowcolor{line-blue} GeoWeaver-5B & 83.33 & \textbf{64.80} {\small (+1.06)} & 58.10 {\small (+2.64)} & \textbf{69.69} {\small (+0.32)} & \textbf{68.98} {\small (+0.98)} \\
            \bottomrule
        \end{tabular}%
    }
    
    \label{tab:general_eval}
    \vspace{-2mm}
\end{table*}

\noindent\textbf{Results on VSI-Bench.}
Table~\ref{tab:vsi-bench} reports detailed results on VSI-Bench. GeoWeaver-5B achieves \textbf{72.8} Avg., outperforming the strongest open-source baseline SpatialStack-5B by \textbf{5.3} points. Scaling to GeoWeaver-10B further improves the score to \textbf{75.05}, yielding a \textbf{7.6}-point gain. GeoWeaver shows clear improvements on geometry-sensitive categories, including object counting, object size estimation, relative direction, and route planning, indicating that token-level geometric grounding provides effective spatial evidence for both object-centric and navigation-oriented reasoning. Compared with proprietary and general-purpose VLMs such as GPT-5, Gemini-3-Pro, and Qwen3-VL-8B, GeoWeaver achieves substantially higher performance, suggesting that the gains come from geometry-grounded visual representations rather than stronger generic multimodal capacity.

\noindent \textbf{Results on ReVSI.}
Table~\ref{tab:revsi_compare} evaluates GeoWeaver on ReVSI, a frame-aware spatial reasoning benchmark where the available visual evidence depends on the input frame budget. This setting stresses the ability to infer geometric relations from partial temporal observations, especially for distance estimation, relative relations, and route planning. GeoWeaver-5B achieves 54.9 average accuracy with 32 frames, and GeoWeaver-10B further improves the score to 57.5, outperforming prior spatial VLMs under their reported native frame settings. The strongest gains appear on relation- and navigation-oriented tasks, showing that token-specific geometric evidence helps the model recover spatial structure from limited frames. The 32-frame setting performs best overall, indicating that effective geometric grounding is more important than simply increasing the number of input frames.

\noindent \textbf{Results on spatial reasoning benchmarks.}
Table~\ref{tab:spatial_reasoning_bench} compares GeoWeaver with representative spatial reasoning methods on six benchmarks. GeoWeaver achieves the best result on \emph{every} benchmark and obtains a new overall score of \textbf{64.8}, improving over the strongest prior method by \textbf{12.2} points. The improvement is especially large on \textbf{SPAR-Bench} (\textbf{69.2}, +26.8), where precise relational understanding and scene-level spatial structure are crucial. We also observe consistent gains on all spatial reasoning benchmarks.

A notable pattern is that GeoWeaver improves not only on benchmarks emphasizing local spatial perception, but also on datasets requiring more compositional and viewpoint-aware reasoning. This broad improvement supports our central claim that the key bottleneck lies in the lack of geometry-grounded visual representations: once visual tokens are grounded with token-adaptive geometric evidence before decoding, the resulting representation becomes more effective across a diverse range of spatial reasoning settings.

\noindent \textbf{General capability evaluation.}
An important question is whether stronger spatial grounding comes at the cost of general multimodal ability. Table~\ref{tab:general_eval} shows that this is not the case. GeoWeaver achieves the best \textbf{overall} score of \textbf{68.98}, improving over SpatialStack-5B by \textbf{0.98} points and also slightly surpassing the base Qwen3.5-4B model. In particular, GeoWeaver performs best on \textbf{Video-MME} (\textbf{64.80}), \textbf{TempCompass} (\textbf{69.69}), and obtains a stronger \textbf{BLINK} score than SpatialStack (58.10 vs.\ 55.46), while keeping \textbf{MMBench} essentially unchanged (83.33 vs.\ 83.42). These results indicate that pre-reasoning geometric grounding does not cause catastrophic forgetting. Instead, it improves spatio-temporal reasoning while preserving the broad multimodal competence of the original model.

\noindent \textbf{Discussion.}
Across all three evaluations, the results consistently support the same conclusion: the benefit of GeoWeaver does not come from simply injecting more geometry into the model, but from using geometry to reshape the visual representation before language reasoning begins. By converting raw multi-level geometric cues into geometry-grounded visual tokens, GeoWeaver yields stronger spatial reasoning performance without sacrificing general capability. This validates our view that geometry should serve as a \emph{representational prerequisite}, rather than merely an auxiliary signal fused later in the reasoning stack.

\subsection{Ablations}

\begin{table*}[t]
\caption{
\textbf{Ablation study on VSI-Bench.}
We analyze GeoWeaver from five aspects: geometry bank construction, geometry bank size, evidence compactness, evidence allocation granularity, and grounding position.
}
\vspace{-2mm}
\centering
\footnotesize
\setlength{\tabcolsep}{2.5pt}
\renewcommand{\arraystretch}{1.08}
\begin{adjustbox}{max width=0.9\linewidth}
\begin{tabular}{l|c|cccc|cccc}
    \toprule
    \multirow{2}{*}{Variant} & \multirow{2}{*}{Avg.}
    & \multicolumn{4}{c|}{\cellcolor{orange!10}\textbf{Numerical Answer}}
    & \multicolumn{4}{c}{\cellcolor{yellow!10}\textbf{Multiple-Choice Answer}} \\
    \cmidrule(lr){3-6}\cmidrule(lr){7-10}
    & & Obj. Count & Abs. Dist. & Obj. Size & Room Size
    & Rel. Dist. & Rel. Dir. & Route Plan & Appr. Order \\
    \midrule

    \rowcolor{line-blue}
    \multicolumn{10}{l}{\textbf{A. Geometry Bank Construction}} \\
    First 12 Layers
    & 67.40 & 81.37 & 50.70 & \textbf{77.00} & 54.89
    & 64.50 & 73.38 & 50.00 & 78.67 \\
    Uniform 12 Layers
    & 69.65 & 81.11 & 53.48 & 76.90 & 57.50
    & 67.50 & \textbf{85.01} & 55.00 & 80.67 \\
    Latter 12 Layers
    & \textbf{72.80} & \textbf{83.20} & \textbf{61.80} & 76.80 & \textbf{62.40}
    & \textbf{72.80} & 83.50 & \textbf{56.30} & \textbf{84.70} \\

    \midrule
    \rowcolor{line-blue}
    \multicolumn{10}{l}{\textbf{B. Geometry Bank Size}} \\
    4 Layers
    & 68.32 & 80.05 & 55.22 & 75.15 & 58.02
    & 67.05 & 79.72 & 51.15 & 81.25 \\
    8 Layers
    & 70.08 & 81.45 & 58.55 & 76.05 & 60.25
    & 69.35 & 81.65 & 53.45 & 82.85 \\
    12 Layers
    & \textbf{72.80} & \textbf{83.20} & \textbf{61.80} & \textbf{76.80} & \textbf{62.40}
    & \textbf{72.80} & \textbf{83.50} & \textbf{56.30} & \textbf{84.70} \\
    16 Layers
    & 72.12 & 82.65 & 60.85 & 76.45 & 61.65
    & 72.15 & 82.75 & 55.05 & 84.05 \\
    All Layers
    & 71.60 & 82.30 & 60.10 & 76.30 & 61.20
    & 71.50 & 82.40 & 54.60 & 83.80 \\

    \midrule
    \rowcolor{line-blue}
    \multicolumn{10}{l}{\textbf{C. Evidence Compactness}} \\
    1 Selected Layer
    & 71.69 & 81.94 & 60.25 & 76.37 & 59.77
    & \textbf{73.50} & 83.27 & 53.75 & 84.67 \\
    2 Selected Layers
    & \textbf{72.80} & \textbf{83.20} & \textbf{61.80} & 76.80 & 62.40
    & 72.80 & \textbf{83.50} & \textbf{56.30} & \textbf{84.70} \\
    3 Selected Layers
    & 72.54 & 83.03 & 61.35 & 76.68 & 63.41
    & 73.00 & 83.18 & 55.00 & 84.67 \\
    4 Selected Layers
    & 72.09 & 82.87 & 60.58 & \textbf{76.92} & \textbf{64.09}
    & 73.00 & 82.79 & 52.50 & 84.00 \\

    \midrule
    \rowcolor{line-blue}
    \multicolumn{10}{l}{\textbf{D. Evidence Allocation Granularity}} \\
    Uniform Averaging
    & 70.35 & 82.10 & 58.40 & 76.30 & 60.10
    & 69.20 & 82.10 & 52.50 & 82.80 \\
    Global Layer Weights
    & 71.28 & 82.50 & 59.80 & 76.50 & 61.10
    & 70.40 & 82.60 & 54.10 & 83.40 \\
    Token-Adaptive Allocation
    & \textbf{72.80} & \textbf{83.20} & \textbf{61.80} & \textbf{76.80} & \textbf{62.40}
    & \textbf{72.80} & \textbf{83.50} & \textbf{56.30} & \textbf{84.70} \\

    \midrule
    \rowcolor{line-blue}
    \multicolumn{10}{l}{\textbf{E. Grounding Position}} \\
    Input-Level Fusion
    & 68.10 & 78.20 & 54.60 & 74.10 & 58.30
    & 66.20 & 79.10 & 49.00 & 80.40 \\
    Decoder-Side Fusion
    & 70.45 & 80.10 & 57.30 & 75.40 & 60.20
    & 68.70 & 81.60 & 52.50 & 82.00 \\
    Pre-Reasoning Grounding
    & \textbf{72.80} & \textbf{83.20} & \textbf{61.80} & \textbf{76.80} & \textbf{62.40}
    & \textbf{72.80} & \textbf{83.50} & \textbf{56.30} & \textbf{84.70} \\

    \bottomrule
\end{tabular}%
\end{adjustbox}
\label{tab:ablation_full}
\vspace{-7mm}
\end{table*}

\noindent \textbf{Effect of geometry layer selection.}
Table~\ref{tab:ablation_full} shows that geometry layer selection substantially affects grounding quality. Using the first 12 layers achieves 67.40 on VSI-Bench, while uniformly sampling 12 layers improves the score to 69.65. The latter 12 layers perform best with 72.80, indicating that deeper geometry features provide more reasoning-relevant structural abstractions. This result supports our motivation that geometry layers are heterogeneous and that spatio-temporal reasoning benefits from a geometry bank composed of more structured evidence.

\paragraph{Effect of geometry bank size.}
We further study how many geometry layers should be included in the evidence bank. As shown in Table~\ref{tab:ablation_full}, using only 4 layers achieves 68.32, and increasing the bank size to 8 layers improves the score to 70.08. The default 12-layer bank performs best with 72.80. However, further enlarging the bank to 16 layers or all layers decreases the score to 72.12 and 71.60, respectively, which suggests that a small bank lacks sufficient geometric diversity, while an overly large bank introduces redundant or weakly related structural cues. Therefore, GeoWeaver benefits from a geometry bank that is expressive enough to cover multi-level spatial evidence but compact enough to avoid unnecessary interference.

\noindent \textbf{Effect of compact geometry grounding.}
We study how many geometry layers should be selected for each visual token. Selecting one layer obtains 71.69, while selecting two layers achieves the best score of 72.80. Increasing the number to three or four layers decreases the score to 72.54 and 72.09, respectively. These results suggest that a single layer lacks complementary evidence, whereas overly broad selection introduces redundant or weakly related cues. Thus, compact evidence selection provides a better balance between geometric diversity and noise suppression.

\noindent \textbf{Effect of evidence allocation granularity.}
We compare different strategies for assigning geometry evidence to visual tokens. Uniform averaging obtains 70.35, and global layer weighting improves the score to 71.28 by learning dataset-level layer preferences. Token-adaptive allocation further increases the score to 72.80, showing that a fixed aggregation strategy is insufficient for diverse spatial regions. This result validates the need to allocate geometry evidence according to token-specific spatial roles.

\noindent \textbf{Effect of grounding position.}
We evaluate where geometry should be introduced in the multimodal pipeline. Input-level fusion achieves 68.10, and decoder-side fusion reaches 70.45. In contrast, pre-reasoning grounding obtains 72.80, showing that geometry is most effective when it shapes visual tokens before language decoding. This comparison supports our central design: visual--geometric alignment should be resolved at the perceptual interface, allowing the language model to reason over geometry-ready representations.

\section{Conclusion}
\label{sec:conclusion}

We present GeoWeaver, a pre-reasoning geometric grounding framework for spatio-temporal vision-language reasoning. GeoWeaver addresses the representational bottleneck of current VLMs by grounding visual tokens with geometric evidence before language decoding. Specifically, it constructs a multi-level geometry bank from a frozen geometry encoder and assigns compact, token-specific geometric evidence to visual tokens through adaptive evidence allocation. This design moves visual--geometric alignment to the perceptual interface and provides the language model with geometry-ready representations for relational and temporal reasoning. Experiments across multiple spatial reasoning benchmarks show that GeoWeaver performs favorably against strong general and spatially enhanced VLMs, while preserving general multimodal capability. Moreover, ablation studies validate the importance of reasoning-relevant geometry layers, token-adaptive evidence allocation, and compact geometric grounding. These results indicate that preparing visual representations before reasoning is a promising direction for geometry-aware multimodal intelligence.
\medskip

{
\small

\bibliographystyle{plain}
\bibliography{./ref}
}

\clearpage
\appendix
\setcounter{table}{0}
\renewcommand{\thetable}{\Alph{table}}

\setcounter{figure}{0}
\renewcommand{\thefigure}{\Alph{figure}}

\setcounter{equation}{0}
\renewcommand{\theequation}{\Alph{section}.\arabic{equation}}

We provide additional details about the training and inference, as well as more experiments on various benchmarks and visualizations in Section~\ref{sec:appendix}. Limitations are analyzed in Section~\ref{sec:limitations}.
\section{Additional Details and Results}
\label{sec:appendix}

This appendix provides additional details for GeoWeaver. We first describe the implementation and training settings of the 5B and 10B variants. We then report additional ReVSI results under different frame budgets and provide qualitative visualizations that motivate token-adaptive geometric evidence allocation.

\subsection{Implementation Details}
\label{sec:app_impl}

\noindent \textbf{Model configuration.}
GeoWeaver is instantiated with two model scales, denoted as GeoWeaver-5B and GeoWeaver-10B. Both variants use a Qwen3.5 as the base model and adopt the native vision encoder to extract semantic visual tokens. For geometric grounding, we use a pretrained VGGT encoder as the geometry branch and keep it frozen throughout training. Following the main paper, we collect the latter half of the VGGT hidden states to construct the multi-level geometry bank and discard non-patch tokens before evidence allocation.

\noindent \textbf{Grounding module.}
Each selected geometry layer is first normalized and spatially aligned to the visual-token resolution through a \(2\times2\) spatial merge. A shared MLP then projects the aligned geometry tokens into the hidden space of the language model. During token-adaptive evidence allocation, each visual token predicts its layer preferences over the geometry bank and selects two geometry layers by default. The selected evidence is aggregated and injected into visual tokens through residual grounding. The output projection of the residual grounding branch is initialized to zero for stable optimization.

\noindent \textbf{Training strategy.}
The 5B variant is trained with full-parameter finetuning. The 10B variant is trained with LoRA-based parameter-efficient finetuning. For LoRA training, we use rank \(r=64\), scaling factor \(\alpha=128\), and dropout \(0.05\). The learning rate and global batch size are kept unchanged between the 5B and 10B training settings. All experiments are conducted on 8 A100 GPUs (80GB). The model is optimized with the standard autoregressive next-token prediction objective, without auxiliary reconstruction, contrastive, or geometry-specific supervision losses.

\begin{table*}[ht]
\caption{
\textbf{Implementation summary of GeoWeaver.}
The 5B variant uses full-parameter finetuning, while the 10B variant uses LoRA-based finetuning. The VGGT geometry encoder remains frozen in both settings.
}
\centering
\small
\setlength{\tabcolsep}{5pt}
\renewcommand{\arraystretch}{1.05}
\begin{tabular}{l|cc}
\toprule
Setting & GeoWeaver-5B & GeoWeaver-10B \\
\midrule
Base model & Qwen3.5 & Qwen3.5 \\
Geometry encoder & Frozen VGGT & Frozen VGGT \\
Geometry layers & Latter half & Latter half \\
Geometry alignment & \(2\times2\) spatial merge & \(2\times2\) spatial merge \\
Selected geometry layers & 2 & 2 \\
Frames & 8 &8 \\
Finetuning strategy & Full-parameter & LoRA \\
LoRA rank \(r\) & -- & 64 \\
LoRA \(\alpha\) & -- & 128 \\
LoRA dropout & -- & 0.05 \\
Learning rate & 1e-5 & 1e-5 \\
Global batch size & 64 & 64 \\
GPUs & 8 A100 & 8 A100 \\
\bottomrule
\end{tabular}

\label{tab:app_impl}
\vspace{-4mm}
\end{table*}
\begin{table*}[t]
\caption{
\textbf{Frame-budget analysis on ReVSI.}
All scores are reported as accuracy (\%). The 16-frame setting does not include room size estimation or route planning. The best result in each column is highlighted in \textbf{bold}.
}
\centering
\resizebox{0.98\textwidth}{!}{%
\begin{tabular}{l|c|c|cccc|ccc}
    \toprule
    \multirow{2}{*}[-0.6ex]{Methods}
    & \multirow{2}{*}[-0.6ex]{Frames}
    & \multirow{2}{*}[-0.6ex]{Avg.}
    & \multicolumn{4}{c|}{\cellcolor{orange!10}Numerical Question}
    & \multicolumn{3}{c}{\cellcolor{yellow!10}Multiple-Choice Question} \\
    \cmidrule(lr){4-7}\cmidrule(lr){8-10}
    & & & Obj. Count & Abs. Dist. & Obj. Size & Room Size
    & Rel. Dist. & Rel. Dir. & Route Plan \\
    \midrule

    \multirow{4}{*}{GeoWeaver-5B}
    & 16 & 54.2
    & 38.6 & 61.1 & 62.8 & --
    & 61.4 & 47.0 & -- \\

    & 32 & 54.9
    & 38.0 & 63.4 & 63.6 & \textbf{54.2}
    & \textbf{65.3} & 46.9 & 53.1 \\

    & 64 & 53.6
    & 38.7 & 62.3 & 62.8 & 52.1
    & 61.6 & 45.7 & 51.6 \\

    & All & 54.1
    & 38.6 & 61.3 & 62.2 & 53.7
    & 63.8 & 46.1 & 53.1 \\

    \midrule

    \multirow{4}{*}{GeoWeaver-10B}
    & 16 & 57.4
    & 40.5 & 64.5 & 69.3 & --
    & 62.1 & 50.5 & -- \\

    & 32 & \textbf{57.5}
    & 42.6 & \textbf{66.8} & \textbf{69.8} & 50.9
    & 64.4 & \textbf{52.4} & \textbf{55.4} \\

    & 64 & 56.9
    & 42.4 & 65.7 & 69.6 & 50.7
    & 65.0 & 51.8 & 53.0 \\

    & All & 57.3
    & \textbf{43.2} & 66.5 & 69.7 & 50.7
    & 65.0 & 51.7 & 54.7 \\
    \bottomrule
\end{tabular}%
}
\label{tab:app_revsi_frame}
\vspace{-4mm}
\end{table*}


\begin{table*}[t]
\caption{
\textbf{Detailed results on SPAR-Bench.}
We follow the original SPAR-Bench table format and report the fine-grained scores of GeoWeaver-5B.  GeoWeaver-5B achieves the best overall score and shows strong gains on medium- and high-level spatial reasoning, especially position matching, distance inference, object spatial relation reasoning, and spatial imagination.
}
\centering
\scriptsize
\setlength{\tabcolsep}{2.5pt}
\renewcommand{\arraystretch}{1.15}
\begin{adjustbox}{max width=\linewidth}
\begin{tabular}{l|c|c|c|cccccccc|c|ccc|c|ccccccccc}
\toprule
\multirow{2}{*}{Methods}
& \multirow{2}{*}{Rank}
& \multirow{2}{*}{Avg.}
& \multirow{2}{*}{Low}
& \multicolumn{8}{c|}{Low-level Spatial Perception}
& \multirow{2}{*}{Medium}
& \multicolumn{3}{c|}{Medium-level Spatial Reasoning}
& \multirow{2}{*}{High}
& \multicolumn{9}{c}{High-level Spatial Reasoning} \\
\cmidrule(lr){5-12}\cmidrule(lr){14-16}\cmidrule(lr){18-26}
& & & 
& Depth-OC & Depth-OC-MV & Depth-OO & Depth-OO-MV
& Dist-OC & Dist-OC-MV & Dist-OO & Dist-OO-MV
& & PosMatch & CamMotion & ViewChgI
& & DistI-OO & DistI-OO-MV & ObjRel-OC-MV & ObjRel-OO & ObjRel-OO-MV
& SpImag-OC & SpImag-OC-MV & SpImag-OO & SpImag-OO-MV \\
\midrule

GPT-4o~\cite{openai_gpt5_systemcard}
& 3 & 38.11 & 36.88
& 51.22 & 44.69 & 21.21 & 19.33
& 41.40 & 44.90 & 36.34 & 35.96
& 26.49 & 27.74 & 25.25 & 19.99
& 43.80 & 65.00 & 64.88 & 44.75 & 50.82 & 43.21
& 29.84 & 32.56 & 27.81 & 35.29 \\

GPT-4.1
& 2 & 41.60 & 41.95
& 48.25 & 42.66 & 20.86 & 23.58
& 58.29 & 52.33 & 48.65 & 40.44
& 44.02 & 59.29 & 28.75 & 22.02
& 42.93 & 71.76 & 67.26 & 46.25 & 54.95 & 41.00
& 30.38 & 29.65 & 20.20 & 24.93 \\

Doubao-1.5
& 1 & 41.74 & 33.24
& 34.89 & 37.22 & 25.24 & 21.40
& 35.04 & 34.99 & 40.58 & 36.57
& 47.36 & 55.73 & 39.00 & 29.47
& 49.49 & 74.71 & 69.64 & 35.75 & 70.33 & 47.65
& 34.95 & 33.14 & 35.76 & 42.86 \\

SpaceR-7B~\cite{ouyang2025spacer}
& 4 & 37.55 & 30.90
& 36.14 & 35.16 & 15.40 & 19.25
& 40.13 & 35.07 & 36.97 & 29.16
& 31.93 & 38.68 & 40.00 & 17.23
& 45.61 & 62.35 & 61.61 & 52.25 & 51.92 & 46.81
& 37.90 & 36.05 & 24.83 & 34.17 \\

InternVL2.5-8B+~\cite{zhang2025flatland}
& -- & 63.25 & \textbf{65.53}
& 81.53 & \textbf{79.22} & \textbf{38.25} & \textbf{35.51}
& \textbf{78.93} & \textbf{79.18} & \textbf{68.13} & \textbf{63.50}
& 63.01 & 78.88 & \textbf{73.00} & \textbf{37.14}
& 61.29 & 86.47 & 79.76 & 64.00 & 69.23 & 59.00
& 47.31 & 50.00 & 42.38 & 53.50 \\

\midrule
\rowcolor{line-blue}
\textbf{GeoWeaver-5B}
& \textbf{1} & \textbf{69.15} & 59.30
& \textbf{82.94} & 76.19 & 25.83 & 19.89
& 77.30 & 77.48 & 66.03 & 48.75
& \textbf{75.13} & \textbf{86.26} & 64.00 & 35.14
& \textbf{80.36} & \textbf{89.12} & \textbf{83.93} & \textbf{91.25} & \textbf{85.44} & \textbf{83.38}
& \textbf{73.39} & \textbf{72.09} & \textbf{66.23} & \textbf{78.43} \\
\bottomrule
\end{tabular}%
\end{adjustbox}

\label{tab:app_sparbench_detail}
\vspace{-4mm}
\end{table*}

\begin{table}[t]
\caption{
\textbf{Category-level comparison on Dyn-Bench~\cite{huang2026thinkingdynamicsmultimodallarge}.}
We aggregate the original Dyn-Bench subcategories into Inter-Object, Object-Scene, and Camera-Object groups following the benchmark~\cite{huang2026thinkingdynamicsmultimodallarge}.  Best and second-best results are \textbf{bolded} and \underline{underlined}, respectively.
}
\centering
\scriptsize
\setlength{\tabcolsep}{3.5pt}
\renewcommand{\arraystretch}{0.95}
\begin{adjustbox}{max width=0.82\linewidth}
\begin{tabular}{l|cccc}
\toprule
Methods & Inter-Object & Object-Scene & Camera-Object & Avg. \\
\midrule
\rowcolor{black!6}\multicolumn{5}{l}{\textbf{Proprietary Models}} \\
GPT-4o~\cite{Gpt-4o} & 46.5 & 63.7 & 46.1 & 50.1 \\
GPT-5~\cite{gpt-5} & 54.7 & 70.2 & \textbf{59.2} & 59.5 \\
Gemini-2.5 Pro~\cite{Gemini2.5} & 56.1 & 64.4 & \underline{55.8} & \underline{59.8} \\

\midrule
\rowcolor{black!6}\multicolumn{5}{l}{\textbf{Open-source General Models}} \\
InternVL3-14B~\cite{InternVL3} & 54.0 & 71.6 & 43.2 & 53.7 \\
InternVL3.5-8B~\cite{InternVL3.5} & 50.8 & 65.3 & 42.7 & 50.3 \\
Qwen2.5-VL-7B~\cite{Qwen2.5-VL} & 50.8 & 69.9 & 42.1 & 51.6 \\
LLaVA-OneVision-1.5-4B~\cite{LLaVA-OneVision-1.5} & 49.8 & 64.6 & 40.9 & 49.9 \\
LLaVA-OneVision-1.5-8B~\cite{LLaVA-OneVision-1.5} & 54.0 & 73.1 & 43.2 & 53.8 \\

\midrule
\rowcolor{black!6}\multicolumn{5}{l}{\textbf{Spatial MLLMs}} \\
SpaceR-7B~\cite{spacer} & 56.2 & 72.7 & 48.6 & 56.5 \\
VST-7B-RL~\cite{vst} & \underline{56.3} & 74.4 & 45.7 & 55.7 \\
Spatial-SSRL-7B~\cite{spatial-ssrl} & 47.5 & 69.4 & 36.7 & 45.9 \\
SpatialLadder-3B~\cite{spatialladder} & 52.0 & 72.7 & 44.0 & 53.6 \\

\midrule
\rowcolor{black!6}\multicolumn{5}{l}{\textbf{Region-level MLLMs}} \\
UniPixel-3B~\cite{Unipixel} & 54.6 & 73.2 & 46.3 & 55.4 \\
UniPixel-7B~\cite{Unipixel} & \textbf{56.4} & \underline{75.4} & 50.1 & 58.1 \\
VideoGLaMM~\cite{VideoGLaMM} & 35.0 & 37.4 & 23.3 & 30.7 \\
Sa2VA-InternVL2.5-8B~\cite{Sa2va} & 49.7 & 66.9 & 40.1 & 49.4 \\
Sa2VA-InternVL3-14B~\cite{Sa2va} & 52.7 & 72.3 & 43.8 & 53.6 \\
Sa2VA-Qwen2.5-VL-7B~\cite{Sa2va,Qwen2.5-VL} & 50.3 & 66.7 & 42.4 & 50.3 \\
Sa2VA-Qwen3-VL-4B~\cite{Sa2va,Qwen3-VL} & 48.8 & 67.5 & 40.8 & 49.8 \\

\midrule
\rowcolor{line-blue}
\textbf{GeoWeaver-5B} & 54.6 & \textbf{77.1} & 52.3 & \textbf{61.3} \\
\bottomrule
\end{tabular}
\end{adjustbox}

\label{tab:app_dynbench_category}
\end{table}

\subsection{Additional Results}
\label{sec:app_revsi}

\textbf{ReVSI} evaluates spatial reasoning under frame-aware settings, where the available visual evidence changes with the input frame budget. This protocol is suitable for evaluating whether a model can recover geometric relations from partial temporal observations. Table~\ref{tab:app_revsi_frame} reports GeoWeaver results under different frame budgets.

The 32-frame setting performs best overall for both model scales. GeoWeaver-5B reaches 54.9 average accuracy, while GeoWeaver-10B further improves the score to 57.5. Increasing the frame budget to 64 or all frames does not consistently improve performance, suggesting that moderate temporal coverage provides sufficient spatial context while reducing redundant visual evidence. The improvement from 5B to 10B further indicates that pre-reasoning geometric grounding benefits from a stronger language reasoning capacity.

\noindent \textbf{SPAR-Bench.}
Table~\ref{tab:app_sparbench_detail} reports the detailed SPAR-Bench results following the original benchmark format. GeoWeaver-5B achieves a 69.15 overall score, outperforming the strongest reported baseline. The improvement is most evident on medium- and high-level reasoning tasks, including position matching, distance inference, object spatial relation reasoning, and spatial imagination. These results indicate that token-adaptive geometric evidence allocation is particularly effective when spatial reasoning requires object-level relations and structured scene understanding.

\paragraph{Dyn-Bench.}
Table~\ref{tab:app_dynbench_category} reports category-level results on Dyn-Bench. GeoWeaver-5B achieves the best overall score of 61.3 among the compared models. The strongest gain appears on Object-Scene reasoning, where GeoWeaver reaches 77.1 and outperforms all listed baselines. This category requires modeling how objects interact with scene structure under dynamic visual changes, which aligns with our design of grounding visual tokens with token-specific geometric evidence. The results suggest that pre-reasoning geometric grounding is particularly effective for scene-anchored spatial-temporal reasoning.
\subsection{Additional Ablation Results}
\label{sec:app_ablation}

We provide detailed ablation results for geometry layer selection and evidence compactness. These results complement the main paper by showing how the construction and use of the geometry bank affect spatio-temporal reasoning.

Table~\ref{tab:app_evidence_compactness} studies how many geometry layers should be selected for each visual token. Selecting two layers achieves the best overall score of 72.80. Using one layer limits complementary evidence, while selecting three or four layers begins to introduce weakly related cues. Aggregating all layers further decreases the score to 71.28, confirming that dense use of the geometry bank weakens token-specific grounding. These results support compact evidence allocation rather than uniform aggregation over all geometry layers.

Table~\ref{tab:app_geo_layers} studies how the construction of the geometry bank affects GeoWeaver. Using the first 12 VGGT layers yields 67.40, while uniformly sampling 12 layers improves the score to 69.65. The latter 12 layers achieve the best overall performance of 72.80, with clear gains on absolute distance, room size, relative distance, route planning, and appearance order. This trend indicates that early geometry layers mainly provide local structural details, whereas later layers contain more reasoning-relevant geometric abstractions. The gap between uniform sampling and latter-layer selection further suggests that simply increasing layer diversity is not sufficient; the geometry bank should be composed of features that are better aligned with spatio-temporal reasoning.

\begin{table*}[t]
 \caption{
    \textbf{Ablation on geometry layer selection.}
    The latter 12 geometry layers provide the best overall performance.
    }
    \centering
    \resizebox{0.98\linewidth}{!}{%
    \begin{tabular}{l|c|cccc|cccc}
        \toprule
        \multirow{2}{*}{Geometry Layers} & \multirow{2}{*}{Avg.}
        & \multicolumn{4}{c|}{\cellcolor{orange!10}Numerical Answer}
        & \multicolumn{4}{c}{\cellcolor{yellow!10}Multiple-Choice Answer} \\
        \cmidrule(lr){3-6}\cmidrule(lr){7-10}
        & & Obj. Count & Abs. Dist. & Obj. Size & Room Size
        & Rel. Dist. & Rel. Dir. & Route Plan & Appr. Order \\
        \midrule
        First 12 Layers & 67.40 & 81.37 & 50.70 & 77.00 & 54.89 & 64.50 & 73.38 & 50.00 & 78.67 \\
        Uniform 12 Layers & 69.65 & 81.11 & 53.48 & 76.90 & 57.50 & 67.50 & \textbf{85.01} & 55.00 & 80.67 \\
        Latter 12 Layers & \textbf{72.80} & \textbf{83.20} & \textbf{61.80} & 76.80 & \textbf{62.40} & \textbf{72.80} & 83.50 & \textbf{56.30} & \textbf{84.70} \\
        \bottomrule
    \end{tabular}%
    }
   
    \label{tab:app_geo_layers}
    \vspace{-4mm}
\end{table*}

\subsubsection{Evidence Compactness}

\begin{figure}[ht]
    \centering
    \includegraphics[width=0.8\linewidth]{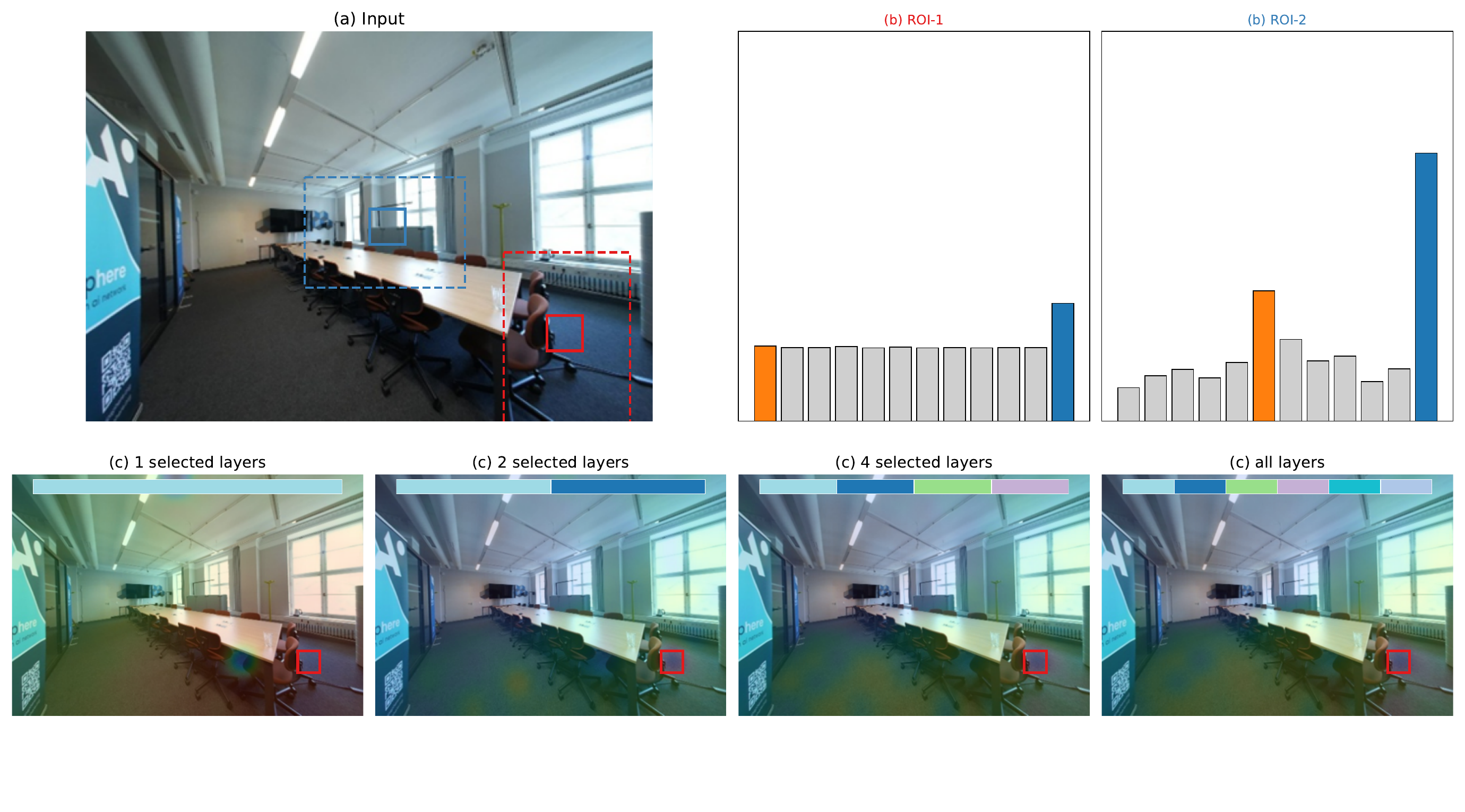}
    \vspace{-4mm}
\caption{
\textbf{Visualization of evidence compactness.}
    We select two representative ROIs and visualize their dense router probability distributions over geometry layers. The distributions concentrate on a small number of layers, suggesting that token-specific geometric evidence is naturally compact. We further compare grounded response maps under different evidence budgets by selecting one, two, four, or all geometry layers for a reference ROI. Too few layers provide incomplete evidence, whereas overly broad selection introduces weakly related geometry and leads to diffuse responses. Compact evidence selection produces a more focused and spatially consistent response, supporting our token-adaptive grounding design.
    }
    \label{fig:evidence_compactness}
    \vspace{-2mm}
\end{figure}

Figure~\ref{fig:evidence_compactness} visualizes why compact geometric evidence is important for token-adaptive grounding. We select two representative ROIs from the same input image and plot their dense router probability distributions over geometry layers. The distributions are concentrated on a small number of layers rather than being uniformly spread across the entire geometry bank, indicating that each visual token only requires a compact subset of geometric evidence. We then fix one ROI as the reference token and construct grounded representations under different evidence budgets, including one, two, four, and all selected layers. The response maps show that using too few layers provides incomplete geometric evidence, while using too many layers introduces weakly related cues and produces more diffuse responses. In contrast, a compact budget yields a more focused and spatially consistent response. The layer indicators shown in each budget further connect the selected geometry layers with the resulting response patterns. This visualization supports the quantitative ablation in Table~\ref{tab:ablation_full}, where compact evidence allocation achieves the best overall performance.

\begin{table*}[t]
 \caption{
    \textbf{Effect of evidence compactness.}
    We vary the number of geometry layers selected for each visual token during token-adaptive evidence allocation. Compact evidence selection achieves the best overall performance, while aggregating all layers introduces weakly related geometric cues and reduces token-specific grounding quality.
    }
    \centering
    \resizebox{0.98\linewidth}{!}{%
    \begin{tabular}{l|c|cccc|cccc}
        \toprule
        \multirow{2}{*}{Selected Layers} & \multirow{2}{*}{Avg.}
        & \multicolumn{4}{c|}{\cellcolor{orange!10}Numerical Answer}
        & \multicolumn{4}{c}{\cellcolor{yellow!10}Multiple-Choice Answer} \\
        \cmidrule(lr){3-6}\cmidrule(lr){7-10}
        & & Obj. Count & Abs. Dist. & Obj. Size & Room Size
        & Rel. Dist. & Rel. Dir. & Route Plan & Appr. Order \\
        \midrule
        1 Layer & 71.69 & 81.94 & 60.25 & 76.37 & 59.77 & \textbf{73.50} & 83.27 & 53.75 & 84.67 \\
        2 Layers & \textbf{72.80} & \textbf{83.20} & \textbf{61.80} & 76.80 & 62.40 & 72.80 & \textbf{83.50} & \textbf{56.30} & \textbf{84.70} \\
        3 Layers & 72.54 & 83.03 & 61.35 & 76.68 & 63.41 & 73.00 & 83.18 & 55.00 & 84.67 \\
        4 Layers & 72.09 & 82.87 & 60.58 & \textbf{76.92} & \textbf{64.09} & 73.00 & 82.79 & 52.50 & 84.00 \\
        All Layers & 71.28 & 82.50 & 59.80 & 76.50 & 61.10 & 70.40 & 82.60 & 54.10 & 83.40 \\
        \bottomrule
    \end{tabular}%
    }
   
    \label{tab:app_evidence_compactness}
    \vspace{-4mm}
\end{table*}

\subsubsection{Geometry Layer Selection}
\begin{figure}[t]
    \centering
    \includegraphics[width=0.7\linewidth]{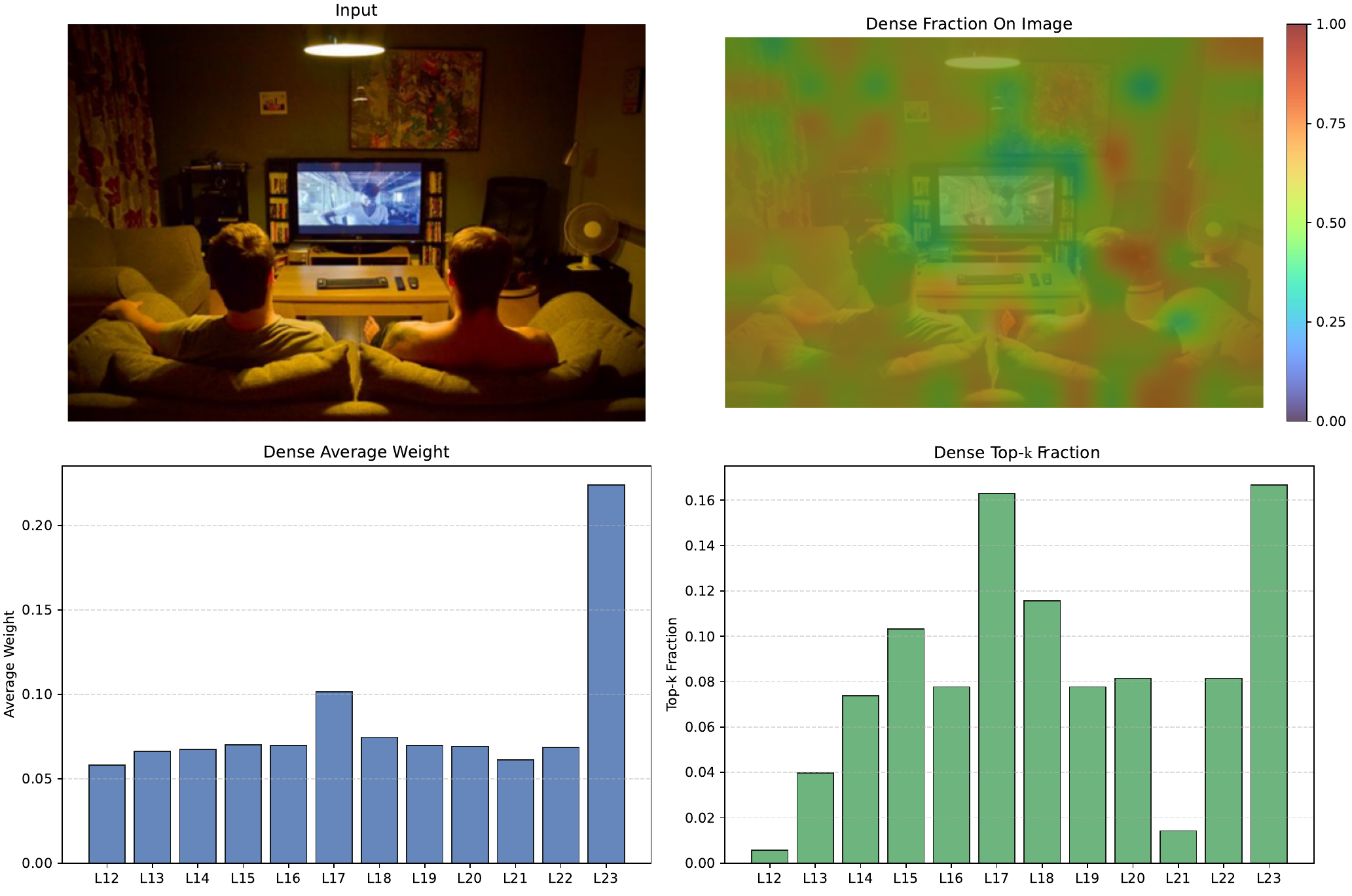}
\caption{
\textbf{Visualization of geometry-layer preference.}
For each spatial location, we compute the average layer index of the top-k geometry layers ranked by dense router scores and overlay it on the input image. Cooler colors indicate a preference for shallower, detail-oriented geometry layers, while warmer colors indicate a preference for deeper, more global geometry layers. The spatially varying pattern shows that different visual tokens require different levels of geometric evidence, supporting our token-adaptive grounding design.
}
    \label{fig:geo_layers}
    \vspace{-2mm}
\end{figure}

To better understand how GeoWeaver allocates geometric evidence, we visualize the spatial distribution of geometry-layer preferences in Figure~\ref{fig:geo_layers}. For each spatial location, we first take the top-k geometry layers according to the dense router scores, compute the average layer index of these candidates, and project the value back to the image plane as a heatmap. Thus, the visualization does not represent the probability of a single layer or the fraction of selected layers. Instead, it reflects the ``center of mass'' of the top candidate geometry layers at each location. Cooler regions indicate that the corresponding visual tokens prefer shallower geometry layers, which usually preserve more local and fine-grained structural details. Warmer regions indicate a stronger preference for deeper geometry layers, which tend to encode more global and semantically organized geometric abstractions. This visualization shows that geometry usage is spatially non-uniform: different image regions rely on different levels of geometric evidence, supporting our token-adaptive evidence allocation design.

\begin{figure*}[t]
    \centering
    \includegraphics[width=0.95\linewidth]{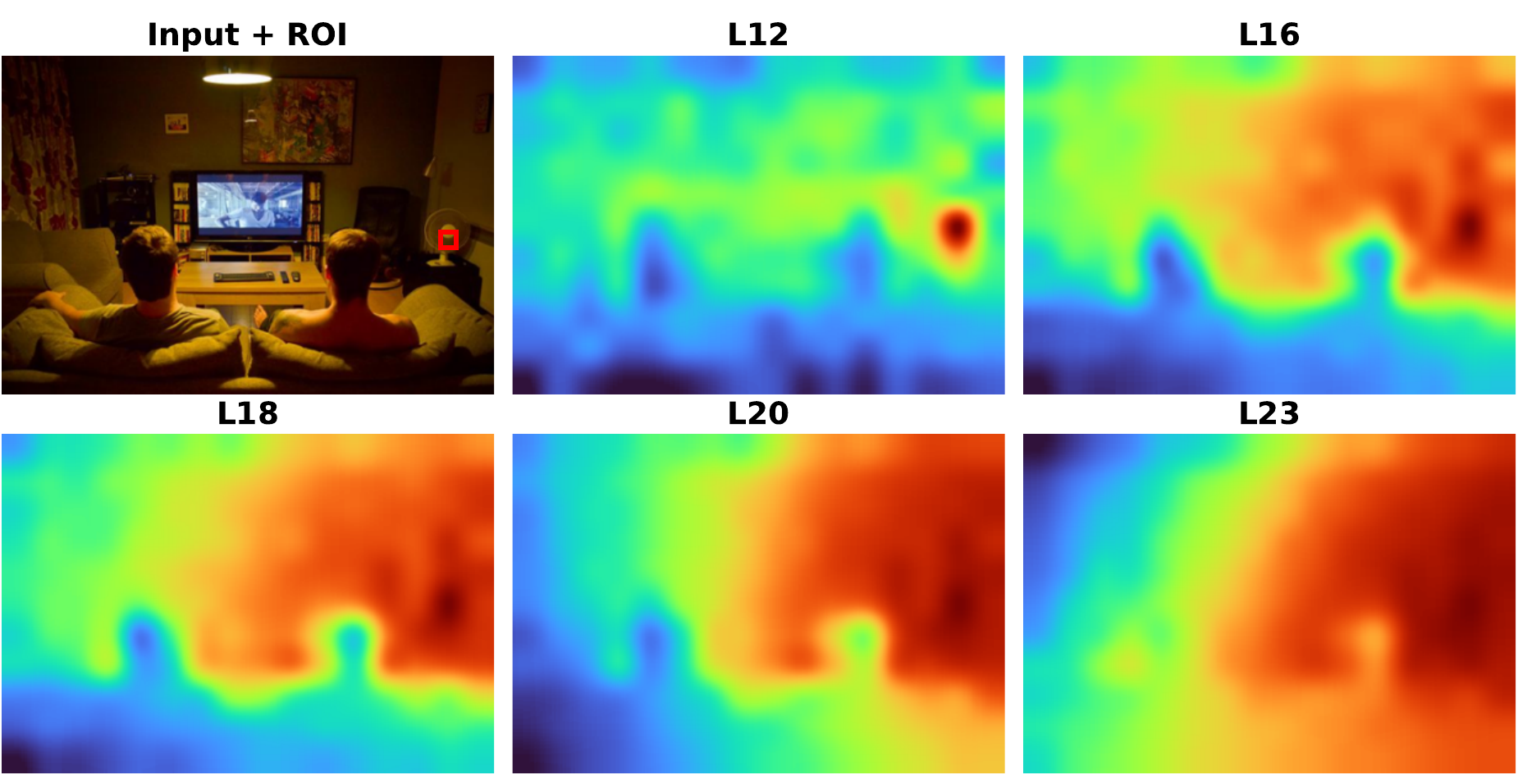}
    \caption{
    \textbf{ROI-to-layer similarity visualization.}
    Given the same ROI, we use the geometry token at the ROI location as a query and compute cosine similarity with all tokens in the same-layer geometry feature map. Different layers exhibit distinct response patterns: earlier layers focus more on local details and boundary-like structures, intermediate layers highlight object-level geometry and nearby context, and later layers capture broader scene layout. This layer-dependent behavior shows that geometry layers encode different types of spatial evidence, motivating the multi-level geometry bank and token-adaptive evidence allocation in GeoWeaver.
    }
    \label{fig:roi_layer_similarity}
\end{figure*}

Figure~\ref{fig:roi_layer_similarity} visualizes how the same ROI interacts with geometry features from different VGGT layers. For each layer, we take the geometry token at the ROI location as the query and compute its cosine similarity with all tokens in the same-layer feature map. The resulting heatmaps show clear layer-dependent responses. Earlier layers, such as L12, produce more localized activations around fine-grained regions and boundary-like structures. Intermediate layers, such as L16 and L18, expand the response to object-level structures and nearby spatial context. Later layers, such as L20 and L23, capture broader scene layout and more global spatial organization. This progression indicates that different geometry layers encode different types of geometric evidence, ranging from local details to global structural abstractions. The visualization supports our use of a multi-level geometry bank and motivates token-adaptive evidence allocation, since different visual tokens may require different geometry layers depending on their spatial roles.

\section{Limitations}
\label{sec:limitations}

GeoWeaver is designed as a modular grounding framework built on top of an external geometry encoder. This design allows the model to benefit from strong geometric priors without requiring additional 3D supervision, but it also means that further improvements may come from stronger or more domain-adaptive geometry encoders. In the current implementation, GeoWeaver uses hidden geometry features as the evidence bank, while explicit geometric states such as depth maps, point maps, or camera poses are not directly modeled. Incorporating these structured 3D representations may further benefit metric-scale reasoning and long-range spatial consistency. 


\end{document}